\documentclass[preprint,review,12pt]{elsarticle}

\usepackage{amssymb}

\usepackage[nodots]{numcompress}

\usepackage{amsfonts} 

\usepackage{color}

\usepackage{graphicx}

\usepackage{times}
\usepackage{epsfig}
\usepackage{amsmath}
\usepackage{amssymb}
\usepackage[caption=false]{subfig}
\usepackage{booktabs}
\usepackage{multirow}

\journal{---}
\newcommand{\figref}[1]{Fig.~\ref{#1}}
\newcommand{\tblref}[1]{Table~\ref{#1}}
\newcommand{\secref}[1]{Section~\ref{#1}}
\renewcommand{\eqref}[1]{Equation~(\ref{#1})}

\newcommand{\ckp}[2]{$CK_{#1}P_{#2}$}

\newcommand{\cfin}{$F_{CK_{X}P_{Y}}$}

\newcommand{\csin}{$SK_{8}P_{8}$}

\makeatletter
\newcommand{\Spvek}[2][r]{%
	\gdef\@VORNE{1}
	\left(\hskip-\arraycolsep%
	\begin{array}{#1}\vekSp@lten{#2}\end{array}%
	\hskip-\arraycolsep\right)}

\def\vekSp@lten#1{\xvekSp@lten#1;vekL@stLine;}
\def\vekL@stLine{vekL@stLine}
\def\xvekSp@lten#1;{\def\temp{#1}%
	\ifx\temp\vekL@stLine
	\else
	\ifnum\@VORNE=1\gdef\@VORNE{0}
	\else\@arraycr\fi%
	#1%
	\expandafter\xvekSp@lten
	\fi}
\makeatother
\begin{document}

	\begin{frontmatter}



		\title{PupilNet v2.0: Convolutional Neural Networks for CPU based real time Robust Pupil Detection}

		

		\author[adrtue]{Wolfgang Fuhl}
		\author[adrtue]{Thiago Santini}
		\author[adrwien]{Gjergji Kasneci}
		\author[adrtue2]{Wolfgang Rosenstiel}
		\author[adrtue]{Enkelejda Kasneci}

		\address[adrtue]{Eberhard Karls University T\"ubingen, Perception Engineering, Germany,72076 T\"ubingen,Sand 14, Tel.: +49 70712970492, wolfgang.fuhl@uni-tuebingen.de, thiago.santini@uni-tuebingen.de, Enkelejda.Kasneci@uni-tuebingen.de}
		\address[adrtue2]{Eberhard Karls University T\"ubingen, Technical Computer Science, Germany,72076 T\"ubingen,Sand 14, Tel.: +49 70712970492, Wolfgang.Rosenstiel@uni-tuebingen.de}
		\address[adrwien]{SCHUFA InnovationLab, SCHUFA Holding AG, Germany, 65201 Wiesbaden, Kormoranweg 5, Tel.: +49 611 92780, gkasneci@googlemail.com}

		\begin{abstract}
Real-time, accurate, and robust pupil detection is an essential prerequisite for
pervasive video-based eye-tracking.  However, automated pupil detection in
real-world scenarios has proven to be an intricate challenge due to fast
illumination changes, pupil occlusion, non-centered and off-axis eye recording,
as well as physiological eye characteristics.
In this paper, we approach this challenge through: I) a convolutional neural
network (CNN) running in real time on a single core, II) a novel computational
intensive two stage CNN for accuracy improvement, and III) a fast propability
distribution based refinement method as a practical alternative to II.
We evaluate the proposed approaches against the state-of-the-art pupil detection
algorithms, improving the detection rate up to $\approx9$\% percent points on average over all data sets.
This evaluation was performed on over 135,000 images: 94,000 images from the
literature, and 41,000 new hand-labeled and challenging images contributed by
this work.
\end{abstract}


	\end{frontmatter}



	\section{Introduction}
\begin{figure}[h]
	\begin{center}
		\subfloat[][\label{a}]{
			\includegraphics[width=.23\columnwidth]{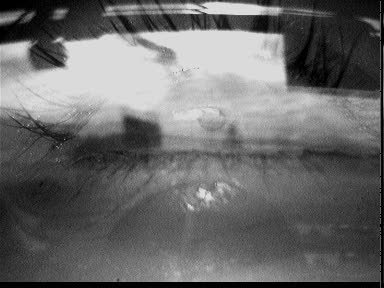}
			\label{fig:reflection}
		}
		\subfloat[][\label{b}]{
			\includegraphics[width=.23\columnwidth]{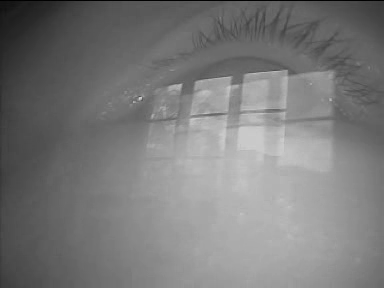}
			\label{fig:reflection2}
		}
		\subfloat[][\label{c}]{
			\includegraphics[width=.23\columnwidth]{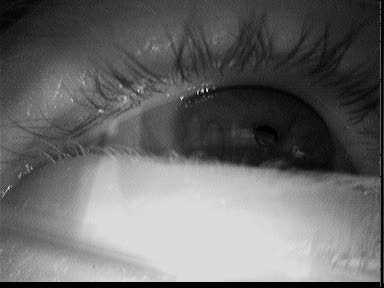}
			\label{fig:dark}
		}
		\subfloat[][\label{d}]{
			\includegraphics[width=.23\columnwidth]{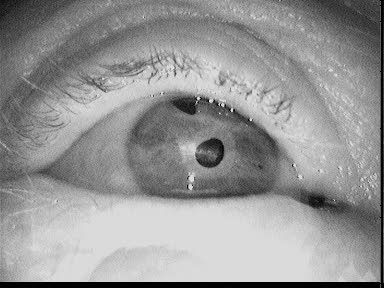}
			\label{fig:physiological}
		}
		\caption{
			Images of typical pupil detection challenges in
			real-world scenarios: (a) and (b) reflections, (c) pupil
			located in dark area, and (d) unexpected physiological structures.
		}
	\end{center}
\end{figure}

For over a century now, the observation and measurement of eye movements have
been employed to gain a comprehensive understanding on how the human oculomotor
and visual perception systems work, providing key insights about cognitive
processes and behavior~\citet{wade2005moving}. Eye-tracking devices are rather
modern tools for the observation of eye movements.
In its early stages, eye tracking was restricted to static activities, such as
reading and image perception~\citet{yarbus1957perception}, due to restrictions
imposed by the eye-tracking system -- e.g., size, weight, cable connections, and
restrictions to the subject itself.
With recent developments in video-based eye-tracking technology, eye tracking
has become an important instrument for cognitive behavior studies in many areas,
ranging from real-time and complex applications (e.g., driving
assistance based on eye-tracking input~\citet{kasneci2013towards} and gaze-based
interaction~\citet{turner2013eye}) to less demanding use cases, such as usability
analysis for web pages~\citet{cowen2002eye}.
Moreover, the future seems to hold promises of pervasive and unobtrusive
video-based eye tracking~\citet{kassner2014pupil}, enabling research and
applications previously only imagined.
Whereas video-based eye tracking has been shown to perform satisfactorily under
laboratory conditions, many studies report the occurrence of difficulties and
low pupil detection rates when these eye trackers are employed for tasks in
natural environments, for instance
driving~\citet{kasneci2013towards,liu2002real,trosterer2014eye} and
shopping~\citet{kasneci2014homonymous}.
The main source of noise in such realistic scenarios is an unreliable pupil
signal, stemming from intricate challenges in the image-based pupil
detection.
A variety of difficulties occurring when using video-based eye trackers, such as
changing illumination, motion blur, and pupil occlusion due to eyelashes, are
summarized in~\citet{schnipke2000trials}.
Rapidly changing illumination conditions arise primarily in tasks where the
subject is moving fast (e.g., while driving) or rotates relative to unequally
distributed light sources, while motion blur can be caused by the image sensor
capturing images during fast eye movements such as saccades.
Furthermore, eyewear (e.g., spectacles and contact lenses) can result in
substantial and varied forms of reflections (\figref{fig:reflection} and
\figref{fig:reflection2}), non-centered or off-axis eye position relative to the
eye-tracker can lead to pupil detection problems, e.g., when the pupil is
surrounded by a dark region (\figref{fig:dark}).
Other difficulties are often posed by physiological eye characteristics, which
may interfere with detection algorithms (\figref{fig:physiological}).
It is worth noticing that such unreliable pupil signals can not only
significantly disturb algorithms for the automatic identification of eye
movements~\citet{santini2016bayesian} but also result in inaccurate gaze estimates.
As a consequence, the data collected in such studies must be post-processed
manually, which is a laborious and time-consuming procedure.
Additionally, this post-processing is impossible for real-time applications that
rely on the pupil monitoring (e.g., driving or surgery assistance).
Therefore, a real-time, accurate, and robust pupil detection is an essential
prerequisite for pervasive video-based eye-tracking.

State-of-the-art pupil detection methods range from relatively simple methods
such as combining thresholding and mass center
estimation~\citet{perez2003precise} to more elaborated methods that attempt to
identify the presence of reflections in the eye image and apply
pupil-detection methods specifically tailored to handle such challenges~\citet{fuhl2015excuse} -- a
comprehensive review is given in \secref{sec:related}.
Despite substantial improvements over earlier methods in real-world scenarios,
these current algorithms still present unsatisfactory detection rates in many
important realistic use cases (as low as 34\%~\citet{fuhl2015excuse}).
However, in this work we show that carefully designed and trained convolutional
neural networks~(CNN)~\citet{domingos2012few,lecun1998gradient}, which rely on
statistical learning rather than hand-crafted heuristics, are a substantial step
forward in the field of automated pupil detection.
CNNs have been shown to reach human-level performance on a multitude of pattern
recognition tasks (e.g., digit recognition~\citet{ciresan2012multi}, image
classification~\citet{krizhevsky2012imagenet}).
These networks attempt to emulate the behavior of the visual processing system
and were designed based on insights from visual perception research.

We propose a dual convolutional neural network pipeline for image-based pupil detection. The first pipeline stage employs a shallow CNN on subregions of a downscaled version of the input image to quickly infer a coarse
estimate of the pupil location. This coarse estimation allows the
second stage to consider only a small region of the original image, thus, mitigating the impact of noise and decreasing computational costs.
The second pipeline stage then samples a small window around the coarse position
estimate and refines the initial estimate by evaluating
subregions derived from this window using a second CNN. We have focused on
robust learning strategies (batch learning) instead of more accurate ones
(stochastic gradient descent)~\citet{lecun2012efficient} due to the fact that an
adaptive approach has to handle noise (e.g., illumination, occlusion,
interference) effectively. The motivation behind the proposed pipeline is (i) to reduce the noise in the coarse estimation of the pupil position, (ii) to reliably detect the exact pupil position from the initial estimate, and (iii) to provide an efficient method that can be run in real-time on hardware architectures without an accessible GPU.

A further contribution of this work is a new hand-labeled data set with more
than 40,000 eye images recorded in real world experiments. This data set
consists of highly challenging eye images containing scattered reflections on
glasses covering the parts or the complete pupil, pupils in dark areas whereby
the contrast to the surrounding area is low, and additional black blobs on the
iris, which may result from eye surgery.
In addition, we propose a method for generating training data in an online-fashion, thus being applicable to the task of pupil center detection in online scenarios. We evaluated the performance of different CNN configurations both in terms of quality and efficiency and report considerable improvements over stat-of-the-art techniques.

	\section{Related work}
\label{sec:related}

During the last two decades, several algorithms have addressed image-based pupil detection.
\citet{perez2003precise} first thresholded the image and compute the mass center of the resulting dark pixels. This process was iteratively repeated in an area
around the previously estimated mass center to determine a new mass center until convergence.
The Starburst algorithm, proposed by \citet{li2005starburst}, first
removed the corneal reflection and then located pupil edge points using an iterative feature-based approach.
Based on the RANSAC algorithm~\citet{fischler1981random}, a
best fitting ellipse is determined, and the final ellipse parameters are selected by applying a model-based optimization.
\citet{long2007high} first down sampled the image and search there for an approximate pupil location. The image area around this location was further processed and a parallelogram-based symmetric mass center algorithm is
applied to locate the pupil center.
In another approach, \citet{lin2010robust} thresholded the image, removed
artifacts by means of morphological operations, and applied inscribed
parallelograms to determine the pupil center.
\citet{keil2010real} first located corneal reflections; afterwards,
the input image was thresholded, the pupil blob was searched in the adjacency of
the corneal reflection, and the centroid of pixels belonging to the blob was
taken as pupil center.
\citet{san2010evaluation} threshold the input image and extract
points in the contour between pupil and iris, which were then fitted to an
ellipse based on the RANSAC method to eliminate possible outliers.
\citet{swirski2012robust} started with
a coarse positioning using Haar-like features. The intensity histogram of the coarse position was clustered using  k-means clustering, followed by a modified RANSAC-based ellipse fit.
The above approaches have shown good detection rates and robustness  in controlled settings, i.e., laboratory conditions.

Three recent methods, SET~\citet{javadi2015set}, ExCuSe~\citet{fuhl2015excuse},
and ElSe~\citet{fuhl2015else}, explicitly address the aforementioned challenges
associated with pupil detection in natural environments.
SET~\citet{javadi2015set} first extracts pupil pixels based on a luminance
threshold. The resulting image is then segmented, and the segment borders are
extracted using a Convex Hull method.  Ellipses are fit to the segments based on
their sinusoidal components, and the ellipse closest to a circle is selected as
pupil.  ExCuSe~\citet{fuhl2015excuse} first analyzes the input image with regard
to reflections based on intensity histograms. Several processing steps based on
edge detectors, morphologic operations, and the Angular Integral Projection
Function are then applied to extract the pupil contour. Finally, an ellipse is
fit to this line using the direct least squares method.
ElSe~\citet{fuhl2015else} is based on the same edge based approach as
ExCuSe~\citet{fuhl2015excuse} with further modifications like improved
morphologic operations and line segment filtering by applying an ellipse fit. In
addition the Angular Integral Projection function is replaced by a weighted blob
detector. Although the latter three methods report substantial improvements over
earlier methods, noise still remains a major issue.  Thus, robust detection,
which is critical in many online real-world applications, remains an open and
challenging problem~\citet{JEMR3657}.

Recent developments in machine learning, especially in the field of
neuronal networks, had a big breakthrough by learning cascaded filter
banks, e.g.,~\citet{krizhevsky2012imagenet,lecun1998gradient}. In particular for
computer vision, there are three main advantages of CNNs when compared to fully
connected neuronal networks.
First, the convolution layers, which are linear filter banks learned
by the CNN can be seen as neuronal network layers with
shared weights.
 In image processing, this is achieved by convolving the weights with the
input layer. As a result, these filters are shift-invariant and applicable to the
entire image (since image statistics are stationary). Furthermore, only the local neighborhood of a location has an influence on the result, i.e, the
spatial information of the response remains through to the neuron position. Each
convolution layer has many of these filters and is usually followed by a pooling
layer. The pooling layer subsamples the data and therefore reduces noise. The
second advantage is the topological structure of a CNN, which arises from
cascading multiple convolution layers. This allows to learn features from lower
level features, which is generally known as deep learning. The third advantage is
the consecutive reduction of the parameters in comparison to a fully connected
neuronal network, which results from the topological structure.

Recent developments in CNNs are multi scale
layers~\citet{gong2014multi,cai2016unified}, the inclusion of transposed
convolutional layers (approximated deconvolution)~\citet{xu2014deep,
long2015fully} and recurrent
CNNs~\citet{liang2015recurrent,pinheiro2014recurrent}.
For example, the multi scale approach
by \citet{gong2014multi} is based on spatial pyramid matching from
\citet{lazebnik2006beyond}. The input image is processed on multiple scales
using ImageNet from \citet{krizhevsky2012imagenet}. The extracted feature
vectors are than feed into a CNN. This approach was evaluated for
classification, recognition, and image retrieval. \citet{cai2016unified} proposed
a CNN architecture with fixed input size capable of handling multiple object
sizes. This multi scale CNN follows the idea of training multiple detectors for
each object size summarized in one CNN. For training they used a multi-task loss
formulation where each label consists of the class and the enclosing bounding
box. Another interesting development is the use of recurrent neuronal
networks~\citet{carpenter1987massively} as convolution layer. The idea behind
recurrent neuronal networks is the usage of information from previous
computations. \citet{liang2015recurrent} proposed the recurrent convolution
layer based approach and showed its applicability for image recognition. For
scene labeling \citet{pinheiro2014recurrent} proposed an architecture, which
corrects itself due to this recurrence information. If it is about detection all
mentioned architectures have to be applied to multiple image locations in a
sliding window approach, due to the fact that each layer reduces the output
size. CNNs with transposed convolutional layers address this problem by
spreading the convolution of one location to multiple positions in the output.
These layers approximate a deconvolution. The architecture with transposed
convolutional layers was first proposed by \citet{long2015fully}. Alternatively
\citet{xu2014deep} trained a CNN to learn real deconvolution filters for image
restoration. The architecture of the net consists of large one dimensional
kernels which represent the separable deconvolution filters.

In our scenario, we want to train a CNN for real time pupil center detection
based on the CPU. Therefore most of the extensions like recurrent neuronal
networks, deconvolution or multi scale networks remain prohibitively expensive.
We used the classical window based approach with coarse and fine positioning to
reduce the computational costs of convolutions. In addition, we propose a fast
direct approach.

	\section{Proposed single- and two-stage CNN approaches}
\begin{figure}[h]
	\begin{center}
		\includegraphics[width=.8\columnwidth]{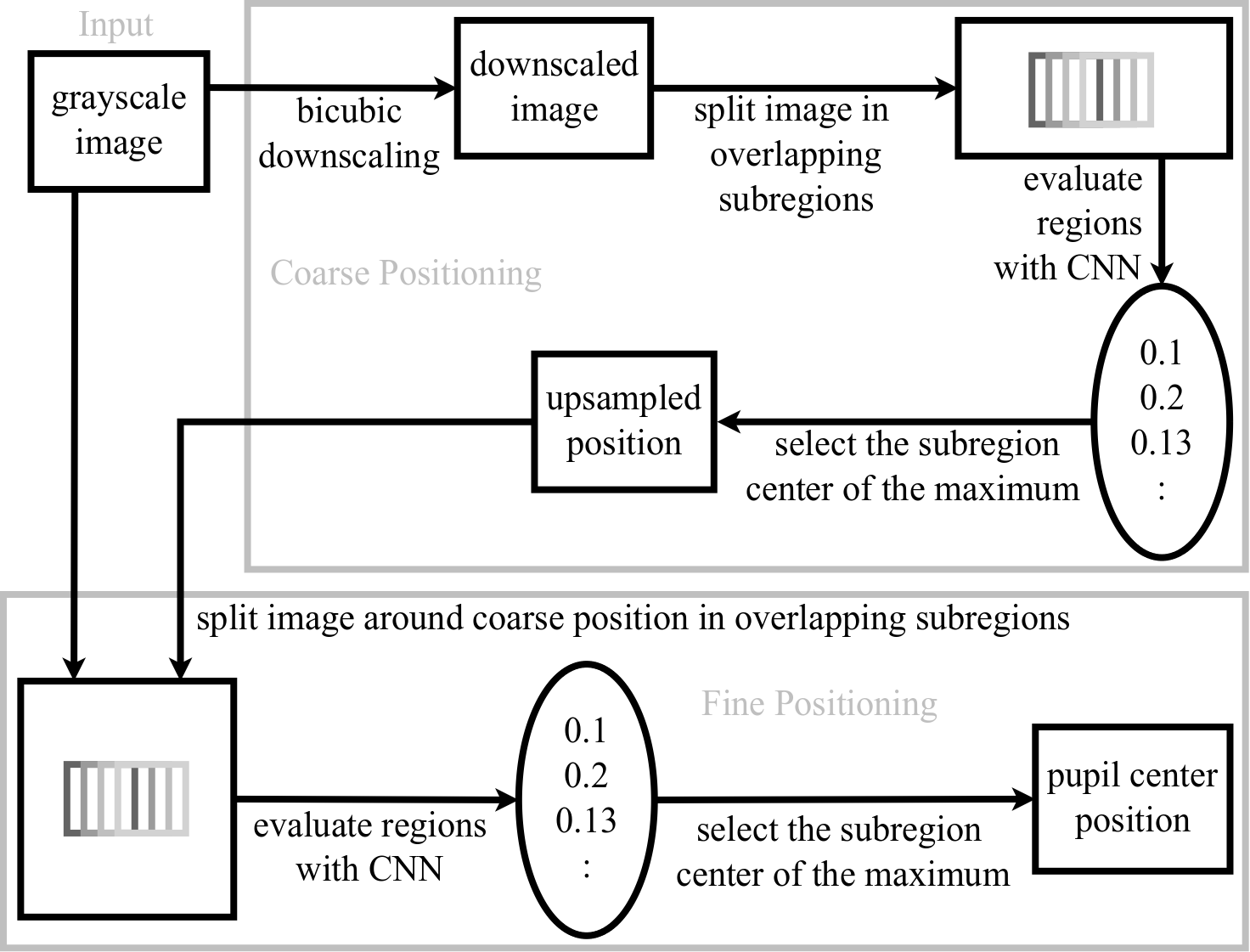}
	\end{center}
	\caption{
		Workflow of the proposed algorithm. First a CNN is employed to estimate
		a coarse pupil position based on subregions from a downscaled version of
		the input image. This position is upsampled to the
		full resolution of the input image (\emph{upsampled position} in the
		workflow diagram). This position is then refined using subregions around
		the coarse estimation in the original input image by a second CNN.
	}
	\label{fig:workflow}
\end{figure}
The overall workflow for the proposed algorithm is shown in
\figref{fig:workflow}.
In the first stage, the image is downscaled and divided into overlapping
subregions. These subregions are evaluated by the first CNN, and the center of
the subregion that evokes the highest CNN response is used as a coarse pupil
position estimate. Afterwards, this initial estimate is fed into the second pipeline stage. In this
stage, subregions surrounding the initial estimate of the pupil position in the
original input image are evaluated using a second CNN. The center of the
subregion that evokes the highest CNN response is chosen as the final pupil
center location. This two-step approach has the advantage that the first step (i.e., coarse
positioning) has to handle less noise because of the bicubic downscaling of the
image and, consequently, involves less computational costs than detecting the
pupil on the complete upscaled image. In the following subsections, we delineate these pipeline stages and their CNN structures in detail, followed by the training procedure employed for each CNN.

\subsection{Overview of all CNNs}
\begin{table}[h]
	\caption{Overview of all evaluated CNN configurations. In row CNN the
	assigned names can be seen. \textbf{C} stands for convolution filter size,
	\textbf{K} is the amount of kernels (or filters), \textbf{D} stands for the
	pooling layer where D comes from down sampling and \textbf{P} stands for the
	amount of perceptron weights in the fully connected layer.}
	\begin{center}
		\begin{tabular}{lllccc|ccc|c|}
			&& & \multicolumn{3}{c}{Layer 1} & \multicolumn{3}{c}{Layer 2} & \multicolumn{1}{c}{}\\ \cline{4-10}
			& & \multicolumn{1}{l|}{\textbf{CNN}} & \textbf{C} & \textbf{K} & \textbf{D}
			& \textbf{C} & \textbf{K} & \textbf{D} & \textbf{P}\\ \cline{3-10}
			&\multirow{3}{*}{Coarse}&\multicolumn{1}{|l|}{\ckp{8}{8}}  & 5 & 8 & 4 & 5 & 8 & - & 8\\
			& &\multicolumn{1}{|l|}{\ckp{8}{16}}  & 5 & 8 & 4 & 5 & 16 & - & 16\\
			& &\multicolumn{1}{|l|}{\ckp{16}{32}} & 5 & 16 & 4 & 5 & 32 & - & 32\\ \cline{3-10}
			& \multirow{1}{*}{Fine}&\multicolumn{1}{|l|}{\cfin}  & 20 & 8 & 5 & 14 & 8 & - &  8\\ \cline{2-10}
			&\multirow{1}{*}{Direct}&\multicolumn{1}{|l|}{\csin{}} & 6 & 8 & 4 & 5 & 8 & - &  8\\
			&\multirow{1}{*}{Fine}&\multicolumn{1}{|l|}{$F_{SK_8P_8}$} & 20 & 8 & 5 & 14 & 8 & - &  8\\ \cline{2-10}
		\end{tabular}
	\end{center}
	\label{tbl:overview}
\end{table}

Table~\ref{tbl:overview} shows an overview of all CNN configurations with their
assigned names. All coarse CNNs follow the core architecture presented
in~\secref{subsec:coarsestage}, and each candidate has a specific number of
filters in the convolution layer as well as perceptron weights in the fully connected
layer. Their names ($CK_{X}P_{Y}$) are prefixed with \underline{C}
(\underline{C}oarse) using X \underline{K}ernels in the first layer and Y
connections to the final \underline{P}erceptron in the fully connected layer.
The second stage CNN (see Figure~\ref{fig:workflow}) is named \underline{F}ine
CNN. The name ($F_{CK_{X}P_{Y}}$) specifies also the assigned coarse positioning
CNN. This CNN is further described in \secref{subsec:finestage}. The last CNN is the
direct pupil center estimation approach \csin{}, where only one
\underline{S}ingle stage is used based on the downsampled image. Those are
described in \secref{subsec:trainingdirect}. However, we evaluated \csin{} also
with the two step approach ($F_{SK_8P_8}$).  

\subsubsection{Coarse positioning CNN (\ckp{8}{8}, \ckp{8}{16}, \ckp{16}{32})}
\label{subsec:coarsestage}
\begin{figure}[h]
	\begin{center}
		\hfill
		\subfloat[][\label{a}]{
			\includegraphics[width=.4\columnwidth]{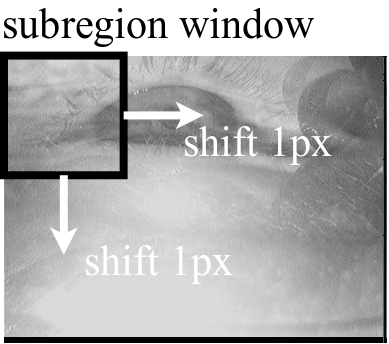}
			\label{fig:splitt}
		}
		\hfill
		\subfloat[][\label{b}]{
			\includegraphics[width=.4\columnwidth]{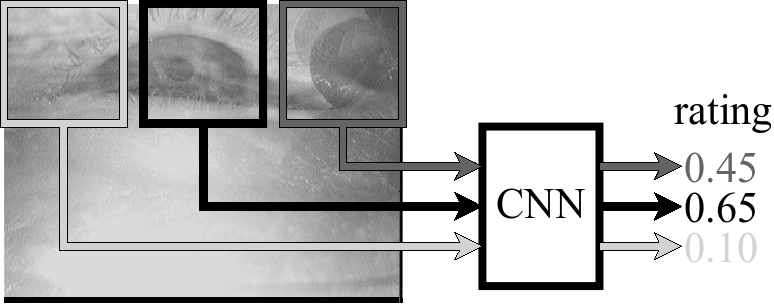}
			\label{fig:cnncoarse}
		}
		\hfill{}
		\caption{
			The downscaled image is divided in subregions of size $24\times24$
			pixels with a stride of one pixel (a), which are then rated by the
			first stage CNN (b).
		}
	\end{center}
\end{figure}
The grayscale input images generated by the mobile eye tracker used in this work
are sized $384\times288$ pixels.
Directly employing CNNs on images of this size would demand a large amount of
resources and, thus, would be computationally expensive, impeding their usage in
state-of-the-art mobile eye trackers. Thus, one of the purposes of the first
stage is to reduce computational costs by providing a coarse estimate that can
in turn be used to reduce the search space of the exact pupil location.
However, the main reason for this step is to reduce noise, which can be induced
by different camera distances, changing sensory systems between head-mounted eye
trackers~\citet{boie1992analysis,dussault2004noise,reibel2003ccd}, movement of the
camera itself, or the usage of uncalibrated cameras (e.g., out of focus,
unbalanced white levels).
To achieve this goal, first the input image is downscaled using a bicubic
interpolation, which employs a third order polynomial in a two dimensional space
to evaluate the resulting values.
In our implementation, we employ a downscaling factor of four times, resulting
in images of $96\times72$ pixels.
Given that these images contain the entire eye, we chose a CNN input size of
$24\times24$ pixels to guarantee that the pupil is fully contained within a
subregion of the downscaled images.
Subregions of the downscaled image are extracted by shifting a $24\times24$
pixels window with a stride of one pixel (see \figref{fig:splitt}) and evaluated
by the CNN, resulting in a rating within the interval [0,1]
(see~\figref{fig:cnncoarse}).

These ratings represent the confidence of the CNN that the pupil center is
within the subregion. Thus, the center of the highest rated subregion is chosen
as the coarse pupil location estimation.
The core architecture of the first stage CNN is summarized in Table~\ref{tbl:overview}.
The first layer is a convolutional layer with filter size $5\times5$ pixels, one
pixel stride, and no padding.
The convolution layer is followed by an average pooling layer with window size
$4\times4$ pixels and four pixels stride. The subsequent stage is an additional
convolution layer with filter size $5\times5$, reducing the size of the feature
map to $1\times1\times8$, which is fed into the last fully connected layer with
depth one.
The last layer can be seen as a single perceptron responsible for yielding
the final rating within the interval [0,1]. The size of the filter in combination
with the pooling size is a trade-off between the information the CNN can hold
and its computational costs. Many small convolution layers would increase the
processing time of the net; in contrast, higher pooling would reduce the
information held by the CNN.

We have evaluated this architecture for different amounts of filters in the
convolutional layers as well as varying the quantity of perceptrons in the fully
connected layer; these values are reported in \secref{sec:eval}.
The main idea behind the selected architecture is that the convolutional layer
learns basic features, such as edges, approximating the pupil structure.
The average pooling layer makes the CNN robust to small translations and
blurring of these features (e.g., due to the initial downscaling of the input
image). The second convolution layer incorporates deeper knowledge on how to combine the
learned features for the coarse detection of the pupil position. The final
perceptron learns a weighting to produce the final rating.

\subsubsection{Fine positioning CNN (\cfin and $F_{SK_8P_8}$)}
\label{subsec:finestage}
Although the first stage yields an accurate pupil position estimate, it lacks
precision due to the inherent error introduced by the downscaling step.
Therefore, it is necessary to refine this estimate.
This refinement could be attempted by applying methods similar to those
described in \secref{sec:related} to a small window around the coarse pupil
position estimate. However, since most of the previously mentioned challenges are not alleviated by
using this small window, we chose to use a second CNN that evaluates subregions
surrounding the coarse estimate in the original image.

The second stage CNN employs the same architecture pattern as the first stage
(i.e., convolution $\Rightarrow$ average pooling $\Rightarrow$ convolution $\Rightarrow$ fully connected) since their motivations are analogous.
Nevertheless, this CNN operates on a larger input resolution to increase
accuracy and precision.
Intuitively, the input image for this CNN would be $96\times96$ pixels: the input
size of the first CNN input ($24\times24$) multiplied by the downscaling factor
($4$).
However, the resulting memory requirement for this size was larger than
available on our test device; as a result, we utilized the closest working size
possible: $89\times89$ pixels.
The size of the other layers were adapted accordingly.
The convolution filters in the first layer were enlarged to $20\time20$ pixels to
compensate for increased noise and motion blur.
The dimension of the pooling window was increased by one pixel,
leading to a decreased input size on the second convolution layer and reduced runtime.

This CNN uses eight convolution filters in the first stage and eight perceptron weights due to the
increased size of the convolution filter and the input region size.
Subregions surrounding the coarse pupil position are extracted based on a window
of size $89\times89$ pixels centered around the coarse estimate, which is
shifted in a radius of $10$ pixels (with a one pixel stride) horizontally and vertically.
Analogously to the first stage, the center of the region with the highest CNN
rating is selected as fine pupil position estimate.
Despite higher computational costs in the second stage, our approach is highly
efficient and can be run on today's conventional mobile eye-tracking systems.

\subsubsection{Direct coarse to fine positioning CNN (\csin{})}
\label{subsec:trainingdirect}

Unfortunately, the fine positioning CNN requires computational capabilities that
are not always found in state-of-the-art embedded systems.
To address this issue, we have developed one additional fine positioning method
for this evaluation that employs a CNN similar to the ones used in the coarse
positioning stage.
However, this CNN uses an input size of $25\times25$ pixels to obtain an even
center.
As a consequence, the first convolution layer was increased to $6 \times 6$ filters.
This method is used as an inexpensive single stage approach (\csin{}) as well as
in combination with the fine positioning CNN ($F_{SK_8P_8}$).

\subsection{CNN training methodology}
\label{subsec:training}
Both CNNs were trained using supervised batch gradient
descent~\citet{lecun2012efficient} with a dynamic learning rate from $10^{-1}$ to $10^{-6}$. The learning rate was dropped after each ten epochs by $10^{-1}$. In the first round we trained for $50$ epochs and selected the best performing CNN on the validation set. This was repeated four times and in each new round the starting learning rate was decreased by a factor of $10^{-1}$. After the last round we did fine tuning by inspecting each iteration additionally. For each round we generated a new training set. The batch size for one iteration was $100$ and all CNNs' weights were initialized using a Gaussian with standard deviation of $0.01$.
While stochastic gradient descent searches for minima in the error plane more
effectively than batch learning when given valid
examples~\citet{heskes1993line,orr1995dynamics}, it is vulnerable to disastrous
hops if given inadequate examples (e.g., due to poor performance of the
traditional algorithm). On the contrary, batch training dilutes this error which is why we have opted for this method.

\subsubsection{Coarse positioning CNN (\ckp{8}{8}, \ckp{8}{16}, \ckp{16}{32})}
\begin{figure}[h]
	\begin{center}
		\includegraphics[width=0.8\textwidth]{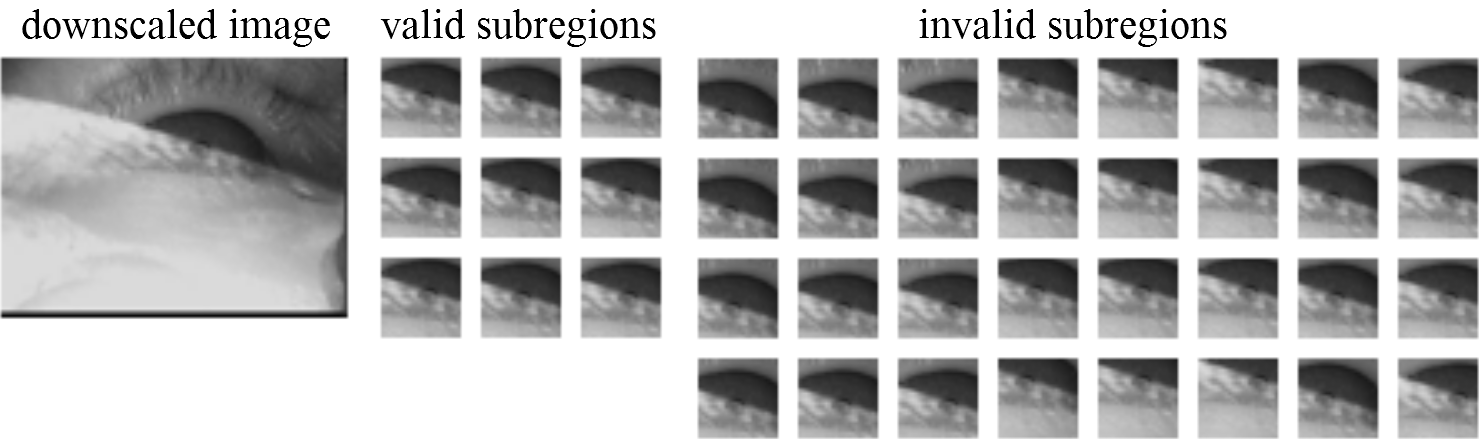}
	\end{center}
	\caption{
		Nine valid (top right) and 32 invalid (bottom) training samples for the coarse
		position CNN extracted from a downscaled input image (top left).
	}
	\label{fig:traindata}
\end{figure}
The coarse position CNN was trained on subregions extracted from the downscaled
input images that fall into two different data classes: containing a valid
($label=1$) or invalid ($label=0$) pupil center.  Training subregions were
extracted by collecting all subregions with center distant up to twelve pixels
from the hand-labeled pupil center. In the first round of training we only used
half of the distance to reduce the amount of invalid examples.  Subregions with
center distant up to one pixel were labeled as valid examples while the
remaining subregions were labeled as invalid examples. As exemplified
by~\figref{fig:traindata}, this procedure results in an unbalanced set of valid
and invalid examples therefore we only used samples on the diagonal (top left to
bottom right) where every second was discarded for the invalid samples. This
reduces the amount of samples per frame. Due to the huge size difference of the
data sets we reduced the amount of samples per set to 20,000 for the first round
and 40,000 for the others. Therefore we picked randomly two thousand images per
data set, created the samples and dropped the overflow. If the data set was to
small we copied the samples to reach the 20,000 or 40,000.

\subsubsection{Fine positioning CNN (\cfin and $F_{SK_8P_8}$)}
The fine positioning CNN (responsible for detecting the exact pupil position) is
trained similarly to the coarse positioning one.
However, we extract only valid subregion up to a distance of three pixels from the hand-labeled
pupil center and selected samples up to a distance of twenty four pixels with a step size of three.  Afterwards the valid examples where again copied to balance the amount of valid and invalid examples. This reduced amount of samples per hand-labeled data ad is to constrain learning time, as well as main memory and storage consumption.

\subsubsection{Direct coarse to fine positioning CNN (\csin{})}
\label{subsec:trainingdirect}

For these CNNs, training and evaluation were performed in an analogous fashion
to the previous ones, with the exception that training samples were generated
from both diagonals (top left to bottom right and top right to bottom left).

\subsection{Fast fine accuracy improvement}
\label{sec:ffai}
The main idea here is to use the response of the CNN surrounding the maximum value to refine the pupil center estimation.
\begin{figure}[h]
	\begin{center}
		\includegraphics[width=0.7\textwidth]{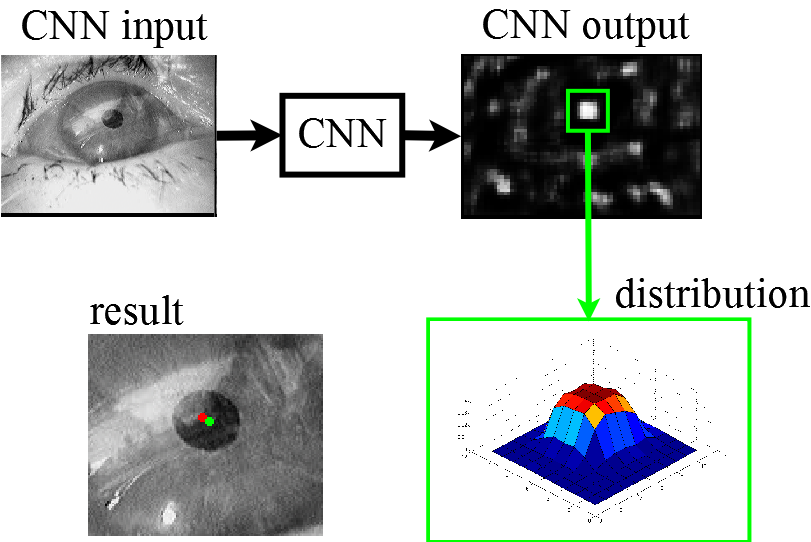}
	\end{center}
	\caption{
		The work flow of the accuracy improvement. On the top left the input image is shown, on the top right is the output of the CNN. For accuracy improvement the surrounding area of the maximum position is converted to a distribution and a shift vector is computed. This distribution is shown in the green box on the bottom right. On the bottom left the maximum position (red dot) and the shifted position (green dot) are shown.
	}
	\label{fig:accimp}
\end{figure}
For a fast accuracy improvement of all CNNs the response surrounding the maximum position is converted into a probability distribution. Such a response of a CNN is shown in figure~\ref{fig:accimp} on the top left. The converted area is surrounded by a green square. In our implementation we used an $7 \times 7$ ($N \times M$) square centered at the maximum position. The resulting distribution is shown in Figure~\ref{fig:accimp} on the bottom right. To convert the response into a distribution each value is divided by the sum of all values in the square (equation~(\ref{eq:distriacc})).
\begin{equation}
D(x,y)=\frac{R(x,y)}{\sum_{i=0}^{N}\sum_{j=0}^{M}R(i,j)}
\label{eq:distriacc}
\end{equation}\\
In equation~(\ref{eq:distriacc}) $D(x,y)$ is the distribution value at location $x,y$ and $R(x,y)$ is the CNN response at location $x,y$. Each value in this distribution is weighted by the displacement vector to the maximum position. The calculation is shown in equation~(\ref{eq:forcevec}).
\begin{equation}
\overrightarrow{SV}=\sum_{i=-\frac{N}{2}}^{\frac{N}{2}}\sum_{j=-\frac{M}{2}}^{\frac{M}{2}}D(\frac{N}{2}+i,\frac{M}{2}+j)*\Spvek{i;j}
\label{eq:forcevec}
\end{equation}\\
In equation~(\ref{eq:forcevec}) $\overrightarrow{SV}$ is the vector shifting the initial maximum position (red dot in figure~\ref{fig:accimp} on the bottom left) to the new more accurate position (green dot in figure~\ref{fig:accimp} on the bottom left). $D(i,j)$ is the result of equation~(\ref{eq:distriacc}) at location $i,j$ and $\Spvek{i;j}$ is the displacement vector to the center.

	\section{Data sets}
\label{sec:newdata}
\begin{figure}[h]
	\begin{center}
		\includegraphics[width=0.8\columnwidth]{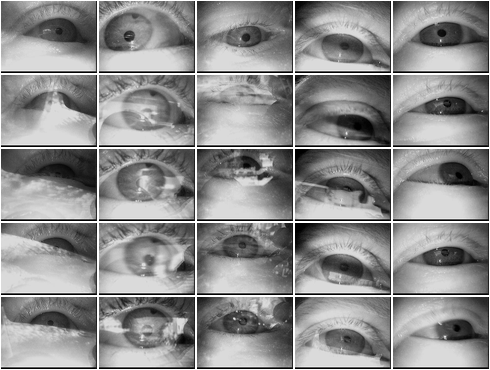}
		\caption{
			Samples from the data sets contributed by this work.
			Each column belongs to a distinct data set.
			The top row includes non-challenging samples, which can be
			considered relatively similar to laboratory conditions and represent
			only a small fraction of each data set.
			The other two rows include challenging samples with artifacts caused
			by the natural environment.
		}
		\label{fig:newdatasets}
	\end{center}
\end{figure}
In this study, we used the data sets provided by
\citet{fuhl2015excuse,fuhl2015else}, complemented by five additional
new hand-labeled data sets contributed by this work. In total, over 135,000 manually labeled eye images were employed for evaluation. Our data sets introduced with this work include 41,217 images collected during driving sessions
in public roads for an experiment \citet{kasneci2013towards} that were not related to pupil
detection and were chosen due the non-satisfactory performance of the
proprietary pupil detection algorithm.
These new data sets include fast changing and adverse illumination, spectacle
reflections, and disruptive physiological eye characteristics (e.g., dark spot
on the iris); samples from these data sets are shown in
\figref{fig:newdatasets}.

	\section{Evaluation}
\label{sec:eval}
\label{sec:expeval}
Training and evaluation figures reported in this paper were obtained on an
Intel\textsuperscript{\textregistered} Core\texttrademark i5-4670 desktop
computer with 8GB RAM. This setup was chosen because it provides a
performance similar to systems that are usually provided by eye-tracker vendors,
thus enabling the actual eye-tracking system to perform other experiments along with the evaluation.
The algorithm was implemented using MATLAB (r2015b) combined with caffe~\citet{jia2014caffe}.
We report our results in terms of the average pupil detection rate as a function
of pixel distance between the algorithmically established and the hand-labeled
pupil center.
Although the ground truth was labeled by experts in eye-tracking research,
imprecision cannot be excluded. Therefore, the results are discussed for a pixel
error of five (i.e., pixel distance between the algorithmically established and
the hand-labeled pupil center), analogously to~\citet{fuhl2015excuse,swirski2012robust}.\\
We performed a per data set cross validation guaranteeing that the CNNs are evaluated on distinct images from those it was trained on. In addition this gives us the advantage for a more detailed comparison between PupilNet and the state-of-the-art algorithms.

\subsection{Coarse positioning}
\begin{figure}[h]
	\begin{center}
		\includegraphics[width=.95\textwidth]{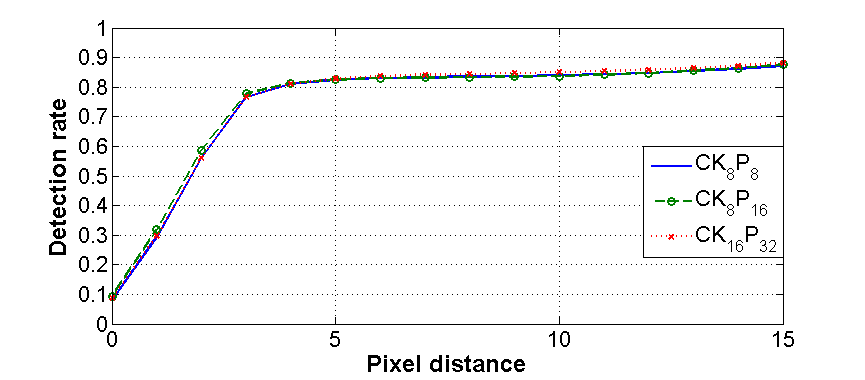}
	\end{center}
	\caption{
		Performance for the evaluated coarse CNNs using the per data set cross validation showing the downscaled error ($*4$ for real error). Each data set is weighted equally meaning that the average result over all data sets is shown independent of their image count.
	}
	\label{fig:evalcoarse}
\end{figure}
We start by evaluating the candidates from~\tblref{tbl:overview} for the coarse
positioning CNN. \figref{fig:evalcoarse} shows the performance of the coarse positioning CNNs
when trained using the per data set cross validation.
As can be seen in figure~\ref{fig:evalcoarse}, the number of filters in the first layer
(\ckp{8}{8}, and \ckp{8}{16}) have only a small impact to the detection rate.
Increasing the amount for both convolutions (\ckp{16}{32}) improves the result slightly but also increases the computational costs (\figref{fig:evalcoarse} shows he average detection rate over all data sets meaning that one percent improvement means an betterment on all data sets). However, it is important to notice
that this is the most expensive parameter in the proposed CNN architecture in
terms of computation time and, thus, further increments must be carefully included.

\subsection{Fine positioning}
\begin{figure}[h]
	\begin{center}
		\includegraphics[width=\textwidth]{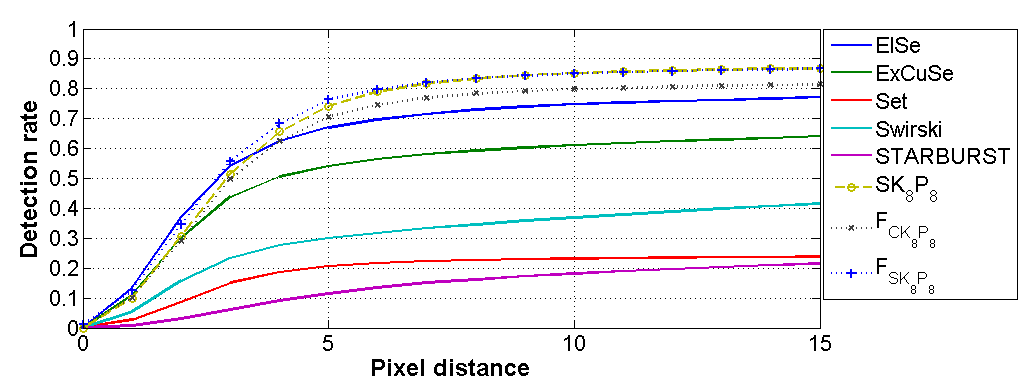}
	\end{center}
	\caption{
		All CNNs were trained and evaluated using the per data set cross validation. The average detection rate over all data sets is shown. The result for \csin{} is with accuracy improvement from section~\ref{sec:ffai}.
	}
	\label{fig:evalfine}
\end{figure}

\setlength{\tabcolsep}{0.2mm}
\renewcommand{\arraystretch}{0.5}
\begin{table}[h!]
	\caption{Five pixel error for the proposed CNNs and the state-of-the-art per data set.}
	\begin{center}
		\begin{tabular}{c|c|c|c|c|c|}
			& ElSe & ExCuSe & \csin{} & \cfin & $F_{SK_{X}P_{Y}}$  \\
			I & \textbf{0.86} & 0.72 & 0.77 & 0.78 & 0.82 \\
			II & 0.65 & 0.40 & \textbf{0.80} & 0.79 & 0.79 \\
			III & 0.64 & 0.38 & 0.62 & 0.60 & \textbf{0.66} \\
			IV & 0.83 & 0.80 & 0.90 & 0.90 & \textbf{0.92} \\
			V & 0.85 & 0.76 & 0.91 & 0.89 & \textbf{0.92} \\
			VI & 0.78 & 0.60 & 0.73 & 0.78 & \textbf{0.79} \\
			VII & 0.60 & 0.49 & 0.73 & \textbf{0.80} & 0.73 \\
			VIII & 0.68 & 0.55 & \textbf{0.84} & 0.83 & 0.81 \\
			IX & \textbf{0.87} & 0.76 & 0.86 & 0.86 & 0.86 \\
			X & 0.79 & 0.79 & 0.80 & 0.78 & \textbf{0.81} \\
			XI & 0.75 & 0.58 & 0.85 & 0.74 & \textbf{0.91} \\
			XII & 0.79 & 0.80 & \textbf{0.87} & 0.85 & 0.85 \\
			XIII & 0.74 & 0.69 & 0.79 & 0.81 & \textbf{0.83} \\
			XIV & 0.84 & 0.68 & 0.91 & 0.94 & \textbf{0.95} \\
			XV & 0.57 & 0.56 & \textbf{0.81} & 0.71 & \textbf{0.81} \\
			XVI & 0.60 & 0.35 & \textbf{0.80} & 0.72 & \textbf{0.80} \\
			XVII & 0.90 & 0.79 & \textbf{0.99} & 0.87 & 0.97 \\
			XVIII & 0.57 & 0.24 & 0.55 & 0.44 & \textbf{0.62} \\
			XIX & 0.33 & 0.23 & 0.34 & 0.20 & \textbf{0.37} \\
			XX & 0.78 & 0.58 & \textbf{0.79} & 0.73 & \textbf{0.79} \\
			XXI & 0.47 & 0.52 & 0.81 & 0.67 & \textbf{0.83} \\
			XXII & 0.53 & 0.26 & 0.50 & 0.52 & \textbf{0.58} \\
			XXIII & \textbf{0.94} & 0.93 & 0.86 & 0.87 & 0.90 \\
			XXIV & 0.53 & 0.46 & 0.46 & \textbf{0.55} & \textbf{0.55} \\
			new I & 0.62 & 0.22 & \textbf{0.69} & 0.56 & \textbf{0.69} \\
			new II & 0.26 & 0.16 & 0.44 & 0.35 & \textbf{0.45} \\
			new III & 0.39 & 0.34 & 0.45 & 0.44 & \textbf{0.49} \\
			new IV & 0.54 & 0.48 & \textbf{0.83} & 0.77 & 0.82 \\
			new V & 0.75 & 0.59 & 0.78 & 0.76 & \textbf{0.81} \\
			\cline{1-6}
		\end{tabular}
	\end{center}
	\label{tbl:eye}
\end{table}

The \cfin was evaluated using all the previously evaluated coarse CNNs (i.e., \ckp{8}{8}, \ckp{8}{16}, and
\ckp{16}{32}). In addition the direct approach \csin{} was evaluated with the accuracy correction from \secref{sec:ffai}. Similarly to the coarse positioning, these were also evaluated
through the per data set cross validation.

As baseline, we evaluated five state-of-the-art algorithms, namely, \emph{ElSe}~\citet{fuhl2015else},
\emph{ExCuSe}~\citet{fuhl2015excuse}, \emph{SET}~\citet{javadi2015set},
\emph{Starburst}~\citet{li2005starburst}, and \citet{swirski2012robust}. The
average performance of the evaluated approaches is shown in~\figref{fig:evalfine}.

As can be seen in the figure, all two-stage CNNs surpass the best performing state-of-the-art approach \emph{ElSe}~\citet{fuhl2015else} by $\approx4$\% and $\approx9$\%. Although the proposed two stage approaches (\cfin and $F_{SK_8P_8}$) reach the best pupil detection rate in average per data set at a pixel error of five, it is worth highlighting the performance of the \csin{} ($\approx7$\% over the state-of-the-art) with its reduced computational costs (runtime of 7ms on a intel i5-4570 3.2GHz single core). This low runtime was reached by only evaluating every second image position in the first step and afterwards extracting the CNN responses on a per pixel level only surrounding the found maximum. The final optimization applied to this region is described in section~\ref{sec:ffai}. In comparison \emph{ElSe} has a runtime of 7ms, \emph{ExCuSe} 6ms and \citet{swirski2012robust} 8ms. \emph{Starburst} and \emph{SET} are not comparable because we used the MATLAB implementation. \cfin has a runtime of 1.2 seconds where in the first step \ckp{8}{8} is used with a runtime of 6ms. Due to the lower accuracy of \ckp{8}{8} in comparison to \csin{} we had to increase the search region of \cfin ($49 \times 49$). This large search region and the high computational costs of \cfin forced us to only evaluate every second image position. For $F_{SK_8P_8}$ we used \csin{} as coarse positioning CNN followed by a fine positioning in a $21 \times 21$ search region. $F_{SK_8P_8}$ has a runtime of 850ms. Due to the architecture of CNNs both approaches \cfin and $F_{SK_8P_8}$ are fully parallelizable with a runtime per patch ($89\times89$) of 2ms. For a finer comparison at a pixel error of five, all results are shown in Table~\ref{tbl:eye}.

	\section{Conclusion}
We presented a naturally motivated pipeline of specifically configured CNNs for
robust pupil detection and showed that it outperforms state-of-the-art
approaches while avoiding high computational costs. For the
evaluation we used over 135,000 hand labeled images -- 41,000 of which were
contributed by this work -- from real-world recordings with artifacts such as
reflections, changing illumination conditions, and occlusions. Specially for
these challenging data sets, the CNNs reported considerably higher detection rates
than state-of-the-art techniques. Looking forward, we are planning to
investigate the applicability of the proposed pipeline to online scenarios,
where continuous adaptation of the parameters is a further challenge. For
further research and usage, data sets, source code for data generation and
training, as well as the trained CNNs will be made available for download.

	\bibliographystyle{elsarticle-num-names}
	\bibliography{references}







\end{document}